\documentclass[10pt]{article} 
\usepackage[preprint]{tmlr}
\usepackage{graphicx}
\usepackage{subcaption}
\usepackage{booktabs}
\usepackage{listings}
\usepackage{xcolor}
\usepackage{tabularx}

\definecolor{codegreen}{rgb}{0.25,0.5,0.35}
\definecolor{codegray}{rgb}{0.5,0.5,0.5}
\definecolor{backcolour}{rgb}{0.97,0.97,0.97}
\definecolor{codebg}{RGB}{245,245,245}
\definecolor{codeblue}{RGB}{40,80,160}
\definecolor{codepurple}{RGB}{160,40,160}

\lstdefinestyle{paperstyle}{
    language=Python,
    backgroundcolor=\color{codebg},
    basicstyle=\ttfamily\footnotesize,
    keywordstyle=\color{codeblue}\bfseries,
    stringstyle=\color{codepurple},
    commentstyle=\color{gray},
    showstringspaces=false,
    breaklines=true,
    frame=single,
    framerule=0.3pt,
    rulecolor=\color{gray},
    xleftmargin=0pt,
    xrightmargin=0pt,
    aboveskip=0pt,
    belowskip=0pt,
    columns=fullflexible,
    keepspaces=true
}

\usepackage{amsmath,amsfonts,bm}

\def\eqref#1{equation~\ref{#1}}

\def\1{\bm{1}}

\DeclareMathAlphabet{\mathsfit}{\encodingdefault}{\sfdefault}{m}{sl}
\SetMathAlphabet{\mathsfit}{bold}{\encodingdefault}{\sfdefault}{bx}{n}

\usepackage{hyperref}
\usepackage{url}

\title{Zero-to-CAD: Agentic Synthesis of Interpretable CAD Programs at Million-Scale Without Real Data}

\author{\name Mohammadmehdi Ataei \email ataei8@gmail.com
      \\
      \addr Autodesk Research
      \AND
      \name Farzaneh Askari \email farzaneh.askari@autodesk.com
      \\
      \addr Autodesk Research
      \AND
      \name Kamal Rahimi Malekshan \email kamal.malekshan@autodesk.com
      \\
      \addr Autodesk Research
      \AND
      \name Pradeep Kumar Jayaraman\textsuperscript{*} \email pradeep.kumar.jayaraman@autodesk.com
      \\
      \addr Autodesk Research
      }

\def\month{MM}  
\def\year{YYYY} 
\def\openreview{\url{https://openreview.net/forum?id=XXXX}} 

\begin{document}

\maketitle
\begingroup
\renewcommand{\thefootnote}{\fnsymbol{footnote}}
\footnotetext[1]{Corresponding author.}
\footnotetext[2]{Dataset and model release: \url{https://huggingface.co/collections/ADSKAILab/zero-to-cad}}
\endgroup

\begin{figure}[h]
\centering
    \includegraphics[width=\textwidth]{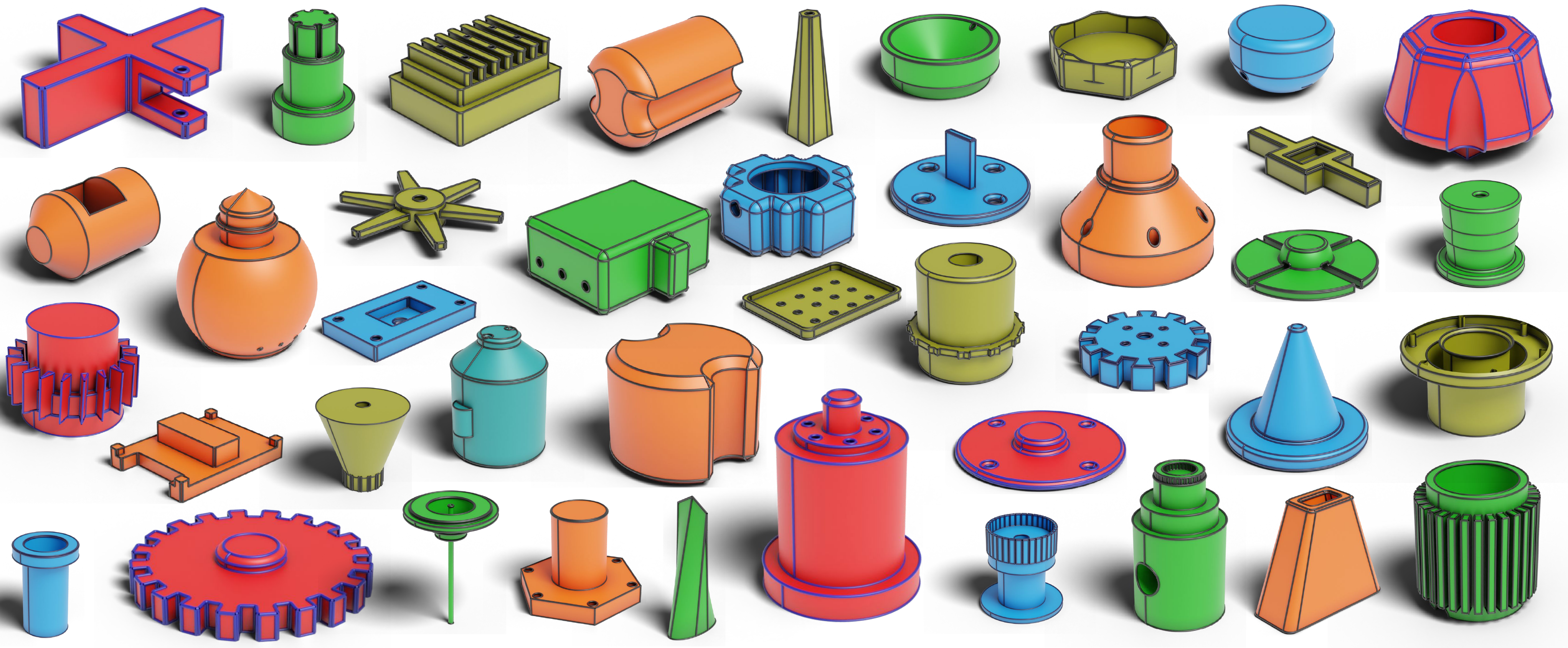}
    \caption{Zero-to-CAD uses an LLM with tool access to generate approximately one million executable CAD construction sequences with interpretable parameters. The examples show diverse mechanical parts, including brackets, housings, gears, and connectors, with fillets, chamfers, holes, and Boolean operations.}
    \label{fig:teaser}
\end{figure}

\begin{abstract}
Computer-Aided Design (CAD) models are defined by their construction history: a parametric recipe that encodes design intent. However, existing large-scale 3D datasets predominantly consist of boundary representations (B-Reps) or meshes, stripping away this critical procedural information. To address this scarcity, we introduce Zero-to-CAD, a scalable framework for synthesizing executable CAD construction sequences. We frame synthesis as an agentic search problem: by embedding a large language model (LLM) within a feedback-driven CAD environment, our system iteratively generates, executes, and validates code using tools and documentation lookup to promote geometric validity and operation diversity. This agentic approach enables the synthesis of approximately one million executable, readable, editable CAD sequences, covering a rich vocabulary of operations beyond sketch-and-extrude workflows. We also release a curated subset of 100,000 high-quality models selected for geometric diversity. To demonstrate the dataset's utility, we fine-tune a vision-language model on our synthetic data to reconstruct editable CAD programs from multi-view images, outperforming strong baselines, including GPT-5.2, and effectively bootstrapping sequence generation capabilities without real construction-history training data. Zero-to-CAD bridges the gap between geometric scale and parametric interpretability, offering a vital resource for the next generation of CAD AI.
\end{abstract}

\section{Introduction}

\begin{table}[h]
\caption{Comparison of CAD datasets that provide construction-sequence information.}
\label{tab:cad-datasets}
\begin{center}
\footnotesize
\setlength{\tabcolsep}{4pt}
\renewcommand{\arraystretch}{1.08}
\begin{tabularx}{\textwidth}{@{}l c c c c >{\raggedright\arraybackslash}X@{}}
\toprule
\textbf{Dataset} & \textbf{Year} & \textbf{Size} & \textbf{Replayable} & \textbf{Readable} & \textbf{Operations and notes} \\
\midrule
DeepCAD~\mbox{\citep{wu2021deepcad}} & 2021 & 178k & Yes & Yes &
\emph{Sketch, extrude.} Human-designed CAD timelines. \\
\addlinespace[2pt]
Fusion 360 Gallery~\mbox{\citep{willis2021fusion}} & 2021 & 8.6k & Yes & Yes &
\emph{Sketch, extrude.} Human-designed CAD timelines. \\
\addlinespace[2pt]
CC3D-Ops~\mbox{\citep{dupont2022cadopsnet}} & 2022 & 37k+ & No & No &
\emph{Extrude/cut, revolve/cut, fillet, chamfer.} Per-face operation-type and step labels; no replayable program. \\
\addlinespace[2pt]
CAD-Recode~\mbox{\citep{rukhovich2025cadrecode}} & 2025 & $\approx$1M & Yes & No &
\emph{Sketch, extrude.} Executable CadQuery code from synthetic sketch-extrude programs. \\
\midrule
\textbf{Zero-to-CAD (ours)} & \textbf{2026} & \textbf{$\approx$1M} & \textbf{Yes} & \textbf{Yes} &
\textbf{Broad operation coverage:} Booleans, fillets, chamfers, lofts, sweeps, shells, and more. Executable, interpretable sequences with a curated subset of 100,000. \\
\bottomrule
\end{tabularx}
\end{center}
\end{table}

Computer-Aided Design (CAD) is the language of physical creation. Unlike meshes or point clouds, a CAD model is often a \textit{program}: a parametric, editable sequence of operations that encodes not just shape, but design intent. This structure allows engineers to modify dimensions, replay histories, and integrate constraints---capabilities that are lost in purely geometric representations.

However, a critical data gap hinders progress in generative CAD. While large-scale datasets such as ABC~\citep{koch2019abc} and Objaverse~\citep{deitke2023objaverse} provide millions of 3D models, they offer only boundary representations (B-Reps) or meshes---geometric snapshots stripped of their parametric history. The few datasets that include construction sequences, such as DeepCAD~\citep{wu2021deepcad} and Fusion 360 Gallery~\citep{willis2021fusion}, are limited in scope, restricted primarily to simple sketch-and-extrude operations that miss the rich vocabulary of real-world design, such as chamfers, fillets, and Boolean operations. Although recent efforts like CAD-Recode~\citep{rukhovich2025cadrecode} generate synthetic code procedurally, these programs often lack the semantic depth and structural diversity found in human-authored models. With most professional design data locked away by proprietary formats and kernel incompatibilities~\citep{heidari2024survey,lin2025cadreconsurvey}, the field lacks a large-scale, diverse source of executable design histories.

We target this need with Zero-to-CAD, a synthesis pipeline that embeds an LLM in a CAD environment with access to tools and documentation. The system proposes candidate construction sequences, executes them in the environment, and uses prompt variability and API-aware checks to broaden part diversity and operation coverage. The goal is not unconstrained procedural scripts, but readable, editable sequences with named parameters, constraints, and references that a human can read and modify. Throughout this paper, ``readable and editable'' means code with explicit named parameters and logical construction steps (see Figure~\ref{fig:example_code}). It is not a user-study-validated measure of comprehensibility; rather, the representational difference from coordinate-chain transpilation methods such as CAD-Recode is directly observable in the released code.

Using this pipeline, we generate and release approximately one million executable construction sequences with complete histories, along with a curated subset of 100,000 chosen for diversity. To our knowledge, this is the first sequence-centric CAD dataset of this scale with broad operation coverage~\cite{koch2019abc,seff2020sketchgraphs,willis2021fusion,wu2021deepcad,dupont2022cadopsnet}. The dataset complements geometry-first datasets by supplying replayable timelines aligned with design intent, and it supports training and evaluation of sequence models. We further demonstrate image-to-sequence modeling from multi-view inputs, which shows a practical path to bootstrap capability without real construction-history data.

\section{Motivation}

We argue that a potential solution to the scarcity of editable, intent-preserving construction histories lies not in collecting more data, but in synthesizing it. While real-world CAD timelines are often unavailable or inconsistent, large language models (LLMs) have absorbed vast amounts of knowledge about object structure and manufacturing processes from textual data. They ``know'' that a bracket needs mounting holes or that a shaft requires a keyway, even if they do not natively produce the precise syntax of a CAD kernel. The challenge is to unlock this latent design knowledge and translate it into valid, executable code.

We address this by framing CAD generation as an \emph{agentic search problem}. Rather than asking a model to generate a perfect program in one shot, we place it in a feedback loop with a CAD interpreter. The agent can write code, execute it, observe errors, read documentation, and inspect the resulting geometry. This grounds the LLM's semantic priors in geometric validity, allowing it to self-correct and produce valid designs that it could never generate in an open-loop setting.

To ensure the resulting dataset spans a wide distribution of shapes and operations, we explicitly design for breadth. We inject randomness into the generation process by varying prompt structures, preventing the model from collapsing into repetitive patterns. In this synthesis regime, exact adherence to a specific prompt is less critical than validity and diversity; we essentially use the LLM to sample the space of plausible mechanical designs, relying on the execution environment to filter out failures. This allows us to cover a vast design space, from simple primitives to complex, multi-feature parts that would be difficult to enumerate manually.

By scaling this process, \emph{we convert compute into data} (i.e., LLM priors about mechanical design are translated into a curated, validated dataset through compute-intensive agentic synthesis). We generate a massive dataset of fully executable, readable, editable CAD sequences from scratch---``Zero-to-CAD''---without relying on real-world CAD files. This synthetic dataset allows us to train smaller, faster, and more specialized models for downstream tasks that perform well at inference time without requiring agentic repair loops or large frontier models. For example, reconstructing editable CAD models from images effectively bootstraps a solution to the sequence generation problem where no construction-history training data previously existed.

\begin{figure*}[h]
    \centering
    \includegraphics[width=\linewidth]{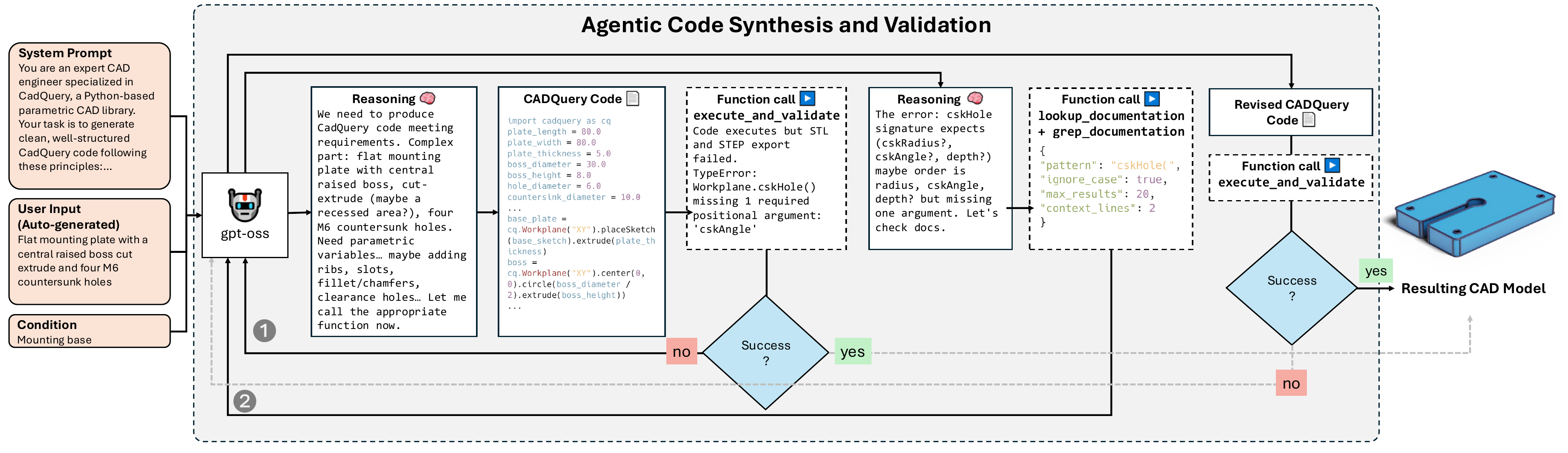}
    \caption{Example of an agentic code synthesis rollout. The LLM generates CadQuery code from a part description, executes it, uses documentation lookup after failures, and revises the code until validation succeeds. See Appendix~\ref{sec:example_code} for the corresponding code.}
    \label{fig:zero_to_cad_architecture}
\end{figure*}

\section{Related Work}

\subsection{CAD Datasets}
Large-scale repositories have primarily focused on boundary representations (B-Reps) and meshes. The ABC dataset~\citep{koch2019abc} collected one million B-Reps, enabling significant advances in geometric deep learning, but explicitly discards construction history, providing only the final B-Rep geometry. SketchGraphs~\citep{seff2020sketchgraphs} offers millions of sketch-and-constraint graphs but does not extend to 3D solid modeling operations. To capture design intent, datasets must include the construction timeline. The Fusion 360 Gallery~\citep{willis2021fusion} provides human-designed sequences but is small (8.6k models) and restricted to sketch-and-extrude operations. DeepCAD~\citep{wu2021deepcad} scales this up synthetically but remains limited to the same narrow operation set. CC3D-Ops~\citep{dupont2022cadopsnet} annotates SolidWorks models with operation types and sequence order, but provides only per-face labels rather than replayable programs.

\subsection{Sketch-and-Extrude Generation}
Traditional CAD modeling relies heavily on 2D sketches lifted into 3D via extrusion or revolution. This paradigm has been adopted by recent generative models trained on datasets such as DeepCAD~\citep{wu2021deepcad} and Fusion 360 Gallery~\citep{willis2021fusion}. DeepCAD treats CAD generation as sequence modeling of sketch-and-extrude commands, and follow-up works have refined this approach: SkexGen~\citep{skexgen} and HNC-CAD~\citep{hnccad} employ hierarchical codebooks to disentangle topology from geometry, while TransCAD~\citep{transcad} conditions generation on point clouds. However, these methods are fundamentally limited by their vocabulary. They operate within a restricted subset of CAD---typically just sketches and extrusions---ignoring critical operations like fillets, chamfers, shells, lofts, and Boolean combinations that define real-world mechanical parts. Furthermore, they depend on the existence of construction history data, which remains scarce. DeepCAD has enabled a family of follow-up works (SkexGen, HNC-CAD, Text2CAD, CAD-Llama, FlexCAD), but its 178k sequences reduce to 114,985 after deduplication~\citep{skexgen} and remain confined to sketch-and-extrude.

\subsection{Direct B-Rep Generation}
A parallel line of research focuses on generating B-Reps directly rather than through construction sequences. SolidGen~\citep{solidgen} pioneered autoregressive B-Rep synthesis by generating faces, edges, and vertices sequentially. BRepGen~\citep{brepgen} introduced a diffusion-based approach using structured latent geometry, representing B-Reps as hierarchical trees. HoLa~\citep{holabrepgen} proposed a holistic latent representation that encodes entire B-Rep models into a unified space, enabling conditional generation from diverse inputs. AutoBrep~\citep{autobrep} unified topology and geometry into a single token sequence for autoregressive generation, achieving state-of-the-art validity and inference speed, while BrepGPT moves to a single-stage autoregressive formulation with a Voronoi Half-Patch representation that also supports conditional generation from modalities such as text and images~\citep{li2025brepgpt}. These methods leverage advances in geometry representation learning: UV-Net~\citep{uvnet} introduced point-grid sampling of parametric surfaces in the UV domain, providing a regular representation invariant to mesh discretization; finite scalar quantization (FSQ)~\citep{fsq} offers an alternative to VQ-VAE for learning discrete codes without codebook collapse. While direct B-Rep methods produce valid geometry, their outputs lack construction histories, limiting downstream editability. Zero-to-CAD complements this by providing the sequences that B-Rep methods lack.

\subsection{Conditional CAD Generation}
Recent work has explored conditioning CAD generation on natural language or images. Text2CAD~\citep{text2cad} generates sequential CAD models from text prompts trained on annotated versions of DeepCAD. CAD-Llama~\citep{cadllama} fine-tunes large language models using structured parametric code representations, achieving high success rates in unconditional generation. FlexCAD~\citep{flexcad} enables controllable generation across CAD construction hierarchies through hierarchy-aware masking of LLM inputs. More recent multimodal systems extend this trend by aligning text, images, and point clouds with CAD command or code representations, including CAD-MLLM, CAD-GPT, and CAD-Coder~\citep{xu2024cad,wang2025cad,doris2026cad}. These approaches demonstrate growing interest in accessible CAD generation interfaces, though they remain constrained by the limited operation coverage and scale of existing sequence datasets.

\subsection{Synthetic Code Generation}
The most relevant precursor, CAD-Recode~\citep{rukhovich2025cadrecode}, generates executable CadQuery code by transpiling synthetic data or procedural trees. However, the resulting scripts often miss the semantic layer of design: they tend to use generic identifiers and hard-coded values rather than the logical parameters and constraints typical of human engineers. In contrast, Zero-to-CAD exploits the semantic knowledge of LLMs to generate designs \textit{ab initio}. This yields interpretable programs with meaningful variable names and a richer operation vocabulary, including Booleans, fillets, and reference geometry, bridging the gap between synthetic execution and human design intent. In Table~\ref{tab:cad-datasets}, we compare CAD datasets that expose construction sequence information by scale, replayability, human readability, and operation coverage. Figure~\ref{fig:dataset_comparison} provides a visual comparison of samples from Zero-to-CAD, ABC, and DeepCAD.

While the individual components---agentic loops, tool use, and LLM code generation---are established techniques, our contribution is integrating them into a robust closed-loop synthesis pipeline that combines two-stage generation, category-conditioned sampling, documentation-grounded repair, and multi-stage validation to enable million-scale dataset creation. The central question we ask is whether LLM priors about plausible mechanical parts can be converted into executable, readable CAD programs without any real construction-history data; our results show they can.

\section{Method}

We employ \texttt{gpt-oss-120b} (served locally under the Apache~2.0 license) in an agentic loop to generate and refine CAD sequences within an interactive environment. The dataset consists entirely of newly synthesized CadQuery programs; no proprietary CAD timelines are extracted or redistributed. Equipped with tools for execution, validation, and documentation lookup, the model iteratively corrects errors and verifies geometric constraints based on runtime feedback.

\subsection{Pipeline Architecture}

A primary challenge in data generation at this scale is building robust, scalable infrastructure. Our synthesis pipeline addresses this through four coordinated components.

\paragraph{LLM Inference Service}
We deploy the LLM on a vLLM-based Ray cluster, enabling efficient multi-turn inference with KV caching. The service exposes an OpenAI-compatible API with function calling, allowing horizontal scaling across GPU workers to support thousands of concurrent rollouts.

\paragraph{Coordinating Node}
A central orchestrator manages independent agentic rollouts, handling load balancing, fault tolerance, and artifact streaming. This architecture decouples generation throughput from model latency, enabling linear scaling with compute resources.

\paragraph{Tool-Equipped Workers}
Each rollout has access to three tools that ground generation in executable reality:

\begin{itemize}
\item \textbf{execute\_and\_validate}: Executes the proposed CadQuery code in an isolated subprocess, performs multi-stage geometric validation, and returns structured feedback including error messages, topology metrics, and export status.
\item \textbf{lookup\_documentation}: Performs TF-IDF-based retrieval over the CadQuery API documentation. We found this lightweight approach sufficient, avoiding the overhead of complex RAG pipelines at scale.
\item \textbf{grep\_documentation}: Provides regex-based pattern matching over documentation for precise syntax lookup when TF-IDF retrieval returns overly broad results.
\end{itemize}

\paragraph{Storage Backend}
Successful sequences and their artifacts (code, STL meshes, STEP files, and metadata) are streamed to cloud storage.

\subsection{Two-Stage Generation Protocol}

The pipeline employs a two-stage generation process that separates the task of deciding \emph{what to build} from \emph{how to build it}. This separation enables controlled diversity across part categories while maintaining geometric validity.

\paragraph{Stage 1: Catalog Generation}
In the first stage, the LLM generates a catalog of part descriptions organized by categories (e.g., ``Bracket'' and ``Gear''). We request descriptions in large batches (typically 200), which encourages diversity as the model uses its context window to avoid repetition within the batch. Acting as a mechanical librarian (Appendix~\ref{sec:system_prompts}), the model produces concise, dimension-free specifications (e.g., ``A mounting bracket with two through-holes'') that are subsequently deduplicated and indexed for downstream generation.

\paragraph{Stage 2: Code Generation from Descriptions}
The second stage takes each description from the catalog and generates executable CadQuery code that implements the described geometry. A part worker receives the description along with a system prompt encoding 19 design principles (see Appendix~\ref{sec:system_prompts} for the full prompt) and, optionally, a reference code snippet that serves as a template. The reference snippet demonstrates coding patterns and geometric techniques but explicitly instructs the model to \emph{adapt} rather than copy: the generated code must implement the new description's geometry, not merely vary parameters of the template. This template-guided generation encourages structured code while preserving diversity across the output space.

\paragraph{Iterative Refinement with Interleaved Reasoning}
Within Stage 2, each part generation follows a multi-turn repair loop that leverages the model's ability to interleave reasoning with tool use, as illustrated in Figure~\ref{fig:zero_to_cad_architecture}. A typical successful generation proceeds as follows: (1) the model reasons about the part description and generates candidate code, (2) invokes an execution tool to test the code, (3) upon receiving error feedback, reasons about the failure mode and decides whether to consult documentation, (4) if needed, queries the API documentation to retrieve relevant information, (5) reasons about how to apply the documentation to fix the error, and (6) generates revised code. This interleaved pattern of reasoning and function calling allows the model to diagnose errors, gather information, and synthesize solutions across multiple turns rather than attempting to solve everything in a single pass. The loop is capped at 10 rollout turns per attempt and 100 attempts per design task. Critically, the system prompt instructs the model to never simplify code to fix problems; instead, it should look up correct syntax and maintain the intended geometric complexity. This prevents the degenerate solution of stripping features until validation passes trivially.

\subsection{Validation Framework}

The validation framework ensures every released sequence is both executable and geometrically sound.

\paragraph{Code Execution Validation}
The proposed code is executed in an isolated subprocess with a timeout. The execution environment extracts the constructed solid and collects initial topology metrics. Execution failures (syntax errors, runtime exceptions, missing imports) are captured with full stack traces for model feedback.

\paragraph{Geometric Validation}
Valid execution does not guarantee valid geometry. The framework performs several geometric checks: topological validity ensures a well-formed solid without self-intersections or degenerate faces; connectivity requires exactly one connected solid, rejecting disconnected bodies that indicate incomplete Boolean unions. Minimum complexity rejects designs with fewer than 7 B-Rep faces to prevent trivial solutions (see Figure~\ref{fig:face_count_dist} for the resulting distribution), and positive volume ensures the result exceeds a minimum threshold, rejecting degenerate zero-volume results.

\paragraph{Export Validation}
Finally, the framework tests export to both STL and STEP formats. Export failures often reveal subtle geometric issues not caught by earlier checks, such as invalid face orientations or unsupported edge configurations. Only designs that pass all three validation stages are accepted into the dataset. This framework guarantees executable, geometrically valid solids but does not enforce full design-for-manufacturability (DFM) rules. DFM constraints are process-dependent (CNC vs.\ casting vs.\ additive manufacturing), require material assumptions, and would demand efficient per-process verifiers usable at the scale of one million parts---well beyond the scope of this work. We do, however, bias generation toward plausible mechanical intent through category-conditioned descriptions and the 19 prompt principles (Appendix~\ref{sec:system_prompts}), which encourage features such as draft angles, accessible faces, and symmetric hole layouts.

\subsection{Diversity Through Structured Categorization}

To prevent mode collapse and ensure broad coverage of mechanical part types, the pipeline employs structured categorization at the description generation stage.

\paragraph{Part Categories}
The catalog is organized into 65 predefined part categories derived from surveys of common mechanical parts in engineering catalogs and manufacturing databases. Categories span structural components (mounting brackets, L-brackets, gusset plates), rotational elements (pulleys, flywheels, cam followers), enclosures (housings, covers, caps), fastening hardware (clamps, retainers, spacers), and many others. Each category receives a target count of descriptions, ensuring balanced representation across the part taxonomy. The LLM generates descriptions within each category, incorporating appropriate features and operations for that part type.

\paragraph{Reference Code Snippets}
For geometrically complex categories such as brackets, housings, and multi-feature mechanical components, we provide reference code snippets as part of the generation prompt. These snippets demonstrate sophisticated CadQuery patterns including sketch composition, Boolean operations, and feature placement. The generation prompt explicitly instructs the model to study the reference code, learn from its structure and patterns, and adapt those techniques to implement the new description by transferring geometric reasoning rather than copying parameters.

\paragraph{Description-Driven Operation Selection}
Rather than sampling operations from fixed distributions, our method derives them directly from the part description. Features like ``reinforcing ribs'' or ``rounded edges'' naturally prompt appropriate operations, such as extrusions or fillets. This semantic grounding ensures coherent designs with broad operation coverage (see Figure~\ref{fig:ops_dist}).

\begin{figure}[h]
  \centering
  \includegraphics[width=0.6\linewidth]{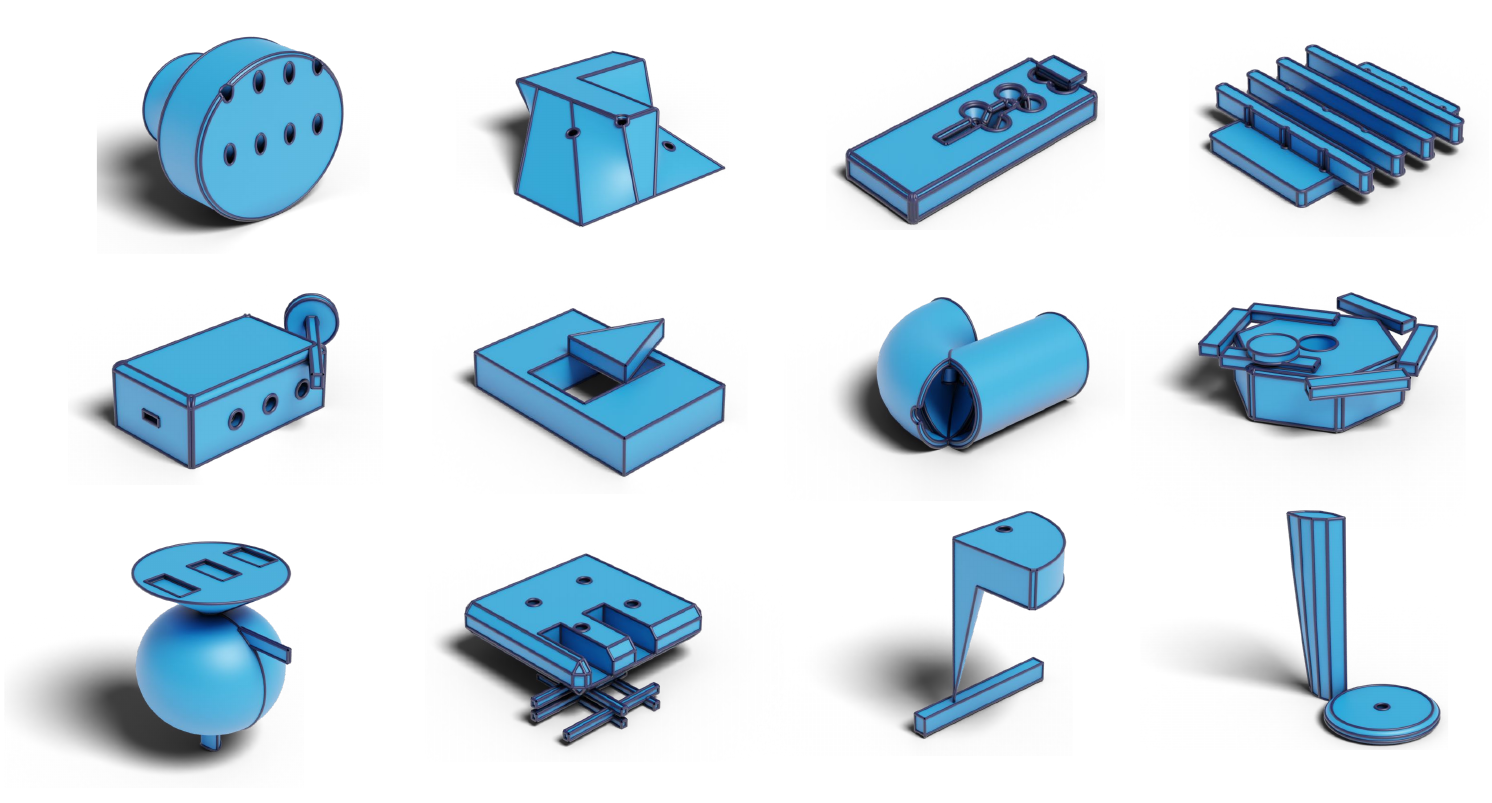}
  \caption{Representative generation failures, including thin features that break connectivity, misplaced holes, self-intersections, scale drift, and locally plausible but globally incoherent primitive compositions.}
  \label{fig:generation_failures}
\end{figure}

\subsection{Computational Resources}

We generated the dataset over approximately one week using opportunistic scheduling on internal idle compute resources. The number of GPUs allocated to LLM inference varied dynamically between 2 and 80 depending on availability. CadQuery execution and function-calling workers ran on CPU nodes, scaling up to 3,000 cores during peak utilization. This elastic approach allowed us to generate approximately one million designs without dedicated infrastructure allocation.

The synthesis pipeline processed about 60 billion input tokens to generate the final dataset. Table~\ref{tab:generation_stats} details key statistics, including token volume, success rates, and function call usage during the agentic repair loops.

\subsection{Curated Subset for Accessibility}

To provide a more accessible entry point for researchers working with limited compute, we release a curated subset of 100,000 selected for diversity. We first compute visual embeddings for each part by averaging DINOv3 features across eight rendered views, then apply k-means clustering to partition the embedding space. From each cluster, we select the nearest-to-centroid exemplar, yielding 100,000 geometrically diverse representatives that span the full distribution of part types. We release both the curated subset and the precomputed DINOv3 embeddings alongside the FAISS index, enabling efficient similarity search over the entire dataset without recomputing features.

\section{Dataset Statistics and Analysis}

The dataset comprises 999,633 executable CAD sequences with full construction histories. Table~\ref{tab:generation_stats} summarizes key generation statistics, with detailed distributions provided in Appendix~\ref{sec:generation_distributions}.

\begin{table}[h]
\caption{Summary of million-scale dataset synthesis. Token counts and tool calls are aggregated across accepted generations and their repair loops.}
\label{tab:generation_stats}
\centering
\small
\setlength{\tabcolsep}{6pt}
\renewcommand{\arraystretch}{1.08}
\begin{tabular}{@{}l r@{}}
\toprule
\textbf{Metric} & \textbf{Value} \\
\midrule
\multicolumn{2}{@{}l}{\textit{Dataset scale}} \\
Total accepted sequences & 999,633 \\
Train / validation / test split & 979,633 / 10,000 / 10,000 \\
\addlinespace[2pt]
\multicolumn{2}{@{}l}{\textit{Token volume}} \\
Total tokens generated & 5.59B \\
Total tokens processed & 60.2B \\
Generated tokens per design & 5,638 mean / 5,089 median \\
\addlinespace[2pt]
\multicolumn{2}{@{}l}{\textit{Repair dynamics}} \\
Function calls per conversation & 4.34 mean \\
First-attempt success rate & 22.3\% \\
Attempts before success & 3.30 mean / 3 median \\
\addlinespace[2pt]
\multicolumn{2}{@{}l}{\textit{Tool usage}} \\
\texttt{execute\_and\_validate} calls & 3.80M \\
\texttt{lookup\_documentation} calls & 375K \\
\texttt{grep\_documentation} calls & 133K \\
\bottomrule
\end{tabular}
\end{table}

\begin{figure*}[h]
  \centering
  \includegraphics[width=0.95\linewidth]{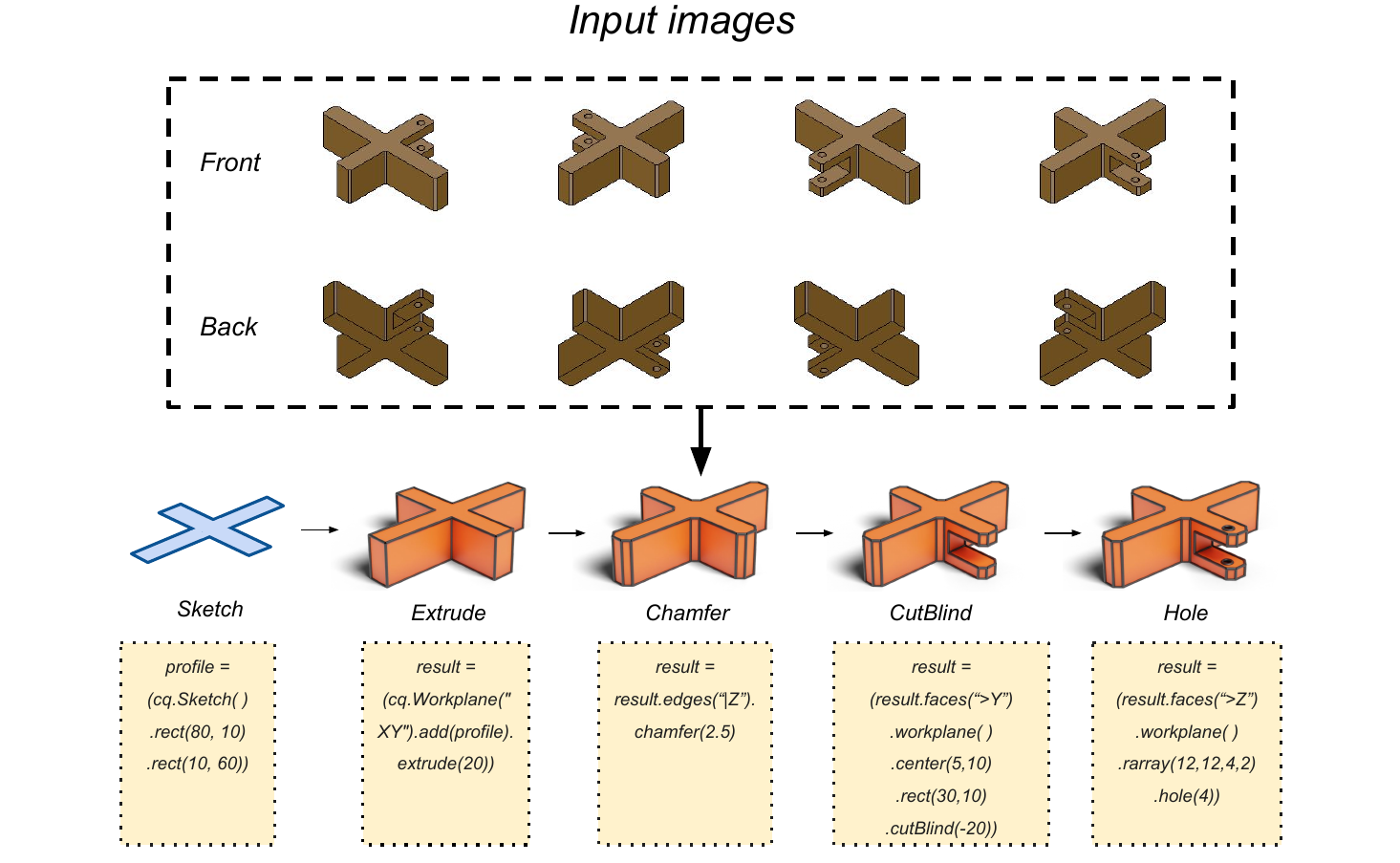}
  \caption{Image-to-Sequence task overview. Given eight rendered views of a CAD model, the fine-tuned VLM generates executable CadQuery code as a sequence of modeling operations.}
  \label{fig:seq}
\end{figure*}

The token distribution is right-skewed, reflecting natural variation in design complexity: simpler primitives require fewer tokens, while multi-operation mechanical parts with extensive parameterization require longer sequences, yielding a mean length of 5,638 tokens. While 22.3\% of designs validate on the first attempt, the majority require iterative refinement. Among function calls, execution validation dominates, confirming its role as the primary driver of refinement.

\paragraph{Generation Success Cases}
Figure~\ref{fig:teaser} shows representative successful generations produced by the synthesis pipeline, illustrating the diversity of part types and operation coverage achievable from executable construction sequences. Each sequence is structured as a logical progression of operations---sketch, extrude, modify---mirroring human design intent and enabling sequential execution.

\paragraph{Generation Failure Modes}
Figure~\ref{fig:generation_failures} highlights characteristic failure modes encountered during synthesis, including thin-wall features that break connectivity, self-intersections, and misplaced holes. These errors typically stem from the LLM's reliance on purely textual reasoning without spatial grounding. While the model constructs locally plausible operations, it struggles to verify global geometric relationships or detect feature intersections that would be obvious in a visual inspection. We do not pursue visual feedback here, as preliminary tests indicated that current models struggle to reliably detect these defects.

\paragraph{Comparison with CAD-Recode}
DeepCAD and CAD-Recode are both constrained to sketch-and-extrude representations; because their vocabulary cannot express fillets, chamfers, shells, lofts, sweeps, or patterns, a direct operation-diversity comparison is not meaningful. We instead compare geometric quality and distributional alignment with ABC. To measure distributional alignment, we compute DINOv2 embeddings for all three datasets and measure alignment with ABC using Fr\'echet distance and $k$-ball coverage (the fraction of ABC shapes with at least one synthetic neighbor within their $k$-th nearest ABC neighbor radius). Zero-to-CAD achieves a lower Fr\'echet distance to ABC (0.164 vs.\ 0.268 for CAD-Recode) and higher coverage at every evaluated $k$ (e.g., 57.2\% vs.\ 45.3\% at $k{=}5$). Geometric quality further separates the two datasets, as shown in Table~\ref{tab:cad_recode_comparison}. Zero-to-CAD's face count distribution closely matches ABC (mean 46.2 vs.\ 50.7), while CAD-Recode's is substantially lower (16.4). Over half of CAD-Recode's parts are disconnected multi-body solids and nearly 13\% fall below our 7-face complexity threshold. Zero-to-CAD enforces single-solid connectivity and minimum face complexity, eliminating both failure modes by construction.

\begin{table}[h]
\caption{Geometric quality and distributional alignment. Lower is better for Fr\'echet distance; higher is better for coverage.}
\label{tab:cad_recode_comparison}
\centering
\small
\setlength{\tabcolsep}{6pt}
\renewcommand{\arraystretch}{1.08}
\begin{tabular}{@{}l c c c@{}}
\toprule
\textbf{Metric} & \textbf{CAD-Recode} & \textbf{Zero-to-CAD} & \textbf{ABC} \\
\midrule
\multicolumn{4}{@{}l}{\textit{Shape complexity}} \\
Mean face count & 16.4 & \textbf{46.2} & 50.7 \\
Median face count & 15 & \textbf{30} & 21 \\
Parts below 7 faces & 12.9\% & \textbf{0\%} & --- \\
Disjoint multi-body solids & 56.3\% & \textbf{0\%} & --- \\
\addlinespace[2pt]
\multicolumn{4}{@{}l}{\textit{Alignment to ABC}} \\
Fr\'echet distance $\downarrow$ & 0.268 & \textbf{0.164} & --- \\
$k$-ball coverage, $k{=}5$ $\uparrow$ & 45.3\% & \textbf{57.2\%} & --- \\
\bottomrule
\end{tabular}
\end{table}

Readability further separates the two datasets. CAD-Recode's transpiled code consists of coordinate chains with no parametric structure, for example:
\par\vspace{4pt}
\noindent\resizebox{\linewidth}{!}{\texttt{r = w0.sketch().segment((\ldots),(\ldots)).segment((\ldots),(\ldots))\ldots .close().finalize().extrude(8)}}
\par\vspace{2pt}
Such sequences are difficult for human engineers to edit manually. Zero-to-CAD programs use named parameters and logical construction order (e.g., \texttt{plate\_thickness}, \texttt{fillet\_radius} in Figure~\ref{fig:example_code}), making design intent explicit and modifications straightforward.

\section{Bootstrapping Experiment}

We present an Image-to-Sequence reconstruction experiment demonstrating that Zero-to-CAD can \emph{bootstrap} sequence-level CAD generation from synthetic supervision (Figure~\ref{fig:seq}). The results below show large gains over the base model and meaningful generalization to human-designed CAD. This experiment is not a controlled dataset-quality ablation against models fine-tuned on CAD-Recode or DeepCAD, because those datasets are largely confined to sketch-and-extrude programs and therefore lack the operation coverage needed to represent many of our target shapes. Its purpose is instead to demonstrate that Zero-to-CAD provides usable supervision at scale and can bootstrap a compact model for sequence generation without any real construction-history data.

\subsection{Experimental Setup}

\paragraph{Task Formulation}
Given eight rendered views of a CAD model at 256$\times$256 resolution (four front-facing and four rear-facing angles), the model must generate executable CadQuery code that reproduces the depicted geometry in a single forward pass. This task requires understanding 3D structure from 2D projections and translating that understanding into parametric code.

\paragraph{Training Data}
We train on the full dataset, split into 979,633 training, 10,000 validation, and 10,000 test samples. Each sample pairs eight rendered 256$\times$256 PNG images with the corresponding CadQuery source code. For out-of-distribution evaluation, we sample 1,000 shapes from the ABC dataset, filtering to retain only models with between 7 and 100 B-Rep faces to exclude trivial and overly complex geometry. Due to API cost constraints, GPT-5.2 models are evaluated on a random subset of 1,000 samples from the Zero-to-CAD test set, while Qwen models are evaluated on the full 10,000-sample Zero-to-CAD test set.

\paragraph{Model}
We fully fine-tune Qwen3-VL-2B-Instruct, a VLM that connects a vision encoder to a language decoder through an MLP adapter.

\paragraph{Training Configuration}
We perform full fine-tuning using distributed data parallelism (DDP) on 16 NVIDIA H100 GPUs; see Appendix~\ref{sec:training_details} for hyperparameters.

\paragraph{Evaluation Metric}
We measure geometric fidelity using voxelized intersection-over-union (IoU) between the generated and ground-truth CAD models. The generated CadQuery code is executed, and both predicted and reference geometries are normalized and voxelized at $64^3$ resolution for volumetric comparison. To account for rotational ambiguity, we rotate the generated shape in increments of 45 degrees and report the maximum IoU. We also report Chamfer distance (CD) as a complementary metric; it shows a consistent pattern with IoU across all models and benchmarks (Table~\ref{tab:iou_summary}). Success rate measures the percentage of generations that produce valid, executable code. Note that success rate alone is an insufficient quality measure: a model that always returns a trivial box achieves 100\% success. IoU and CD together are therefore the primary indicators of geometric fidelity.

\paragraph{Baselines}
We compare against: (1) the base Qwen3-VL-2B-Instruct model without fine-tuning, establishing zero-shot capability of vision-language models on this task, and (2) GPT-5.2 at two reasoning levels (High and Medium), representing state-of-the-art proprietary models. The inference system prompts are very similar (see Appendix~\ref{sec:system_prompts}), with zero-shot models (base Qwen and GPT-5.2) requiring only explicit output-format instructions. These two controls serve the bootstrapping goal: the base-vs.-fine-tuned comparison isolates the effect of Zero-to-CAD supervision, while the GPT-5.2 comparison tests whether specialized training on synthetic data outperforms general-purpose reasoning at inference time. Fine-tuning on DeepCAD or CAD-Recode would not be an informative control because those datasets cannot express the operations (fillets, chamfers, shells, lofts) needed to represent the majority of ABC shapes, making most reconstruction targets unrepresentable. Conversely, ABC cannot supervise the image-to-sequence task because it provides no construction histories or executable programs---this is precisely the gap Zero-to-CAD addresses.

\subsection{Quantitative Results}

Table~\ref{tab:iou_summary} presents IoU metrics across models evaluated on both Zero-to-CAD test samples and the ABC dataset, the latter serving as an out-of-distribution generalization test.

\begin{table}[h]
\caption{Image-to-Sequence reconstruction results. Success measures executable code generation; IoU and Chamfer distance (CD) are computed over successful samples. $^*$Evaluated on 1,000 samples from the Zero-to-CAD test set because of API cost constraints.}
\label{tab:iou_summary}
\centering
\small
\setlength{\tabcolsep}{3.2pt}
\renewcommand{\arraystretch}{1.08}
\begin{tabular}{@{}l c cccc cccc@{}}
\toprule
\textbf{Benchmark / Model} & \textbf{Succ.} 
& \multicolumn{4}{c}{\textbf{IoU $\uparrow$}} 
& \multicolumn{4}{c}{\textbf{CD $\downarrow$}} \\
\cmidrule(lr){3-6}\cmidrule(l){7-10}
& \textbf{(\%)} & \textbf{Mean} & \textbf{Median} & \textbf{P75} & \textbf{P90}
& \textbf{Mean} & \textbf{Median} & \textbf{P75} & \textbf{P90} \\
\midrule
\multicolumn{10}{@{}l}{\textit{Zero-to-CAD test set}} \\
Qwen3-VL-2B (fine-tuned) & \textbf{82.1}
& \textbf{0.747} & \textbf{0.847} & \textbf{0.975} & \textbf{0.999}
& \textbf{0.0143} & \textbf{0.001} & \textbf{0.007} & \textbf{0.03} \\
Qwen3-VL-2B (base) & 6.6
& 0.184 & 0.129 & 0.256 & 0.405
& 0.31 & 0.25 & 0.47 & 0.70 \\
GPT-5.2 High$^*$ & 72.2
& 0.485 & 0.479 & 0.658 & 0.795
& 0.028 & 0.01 & 0.02 & 0.06 \\
GPT-5.2 Medium$^*$ & 71.1
& 0.495 & 0.499 & 0.678 & 0.801
& 0.029 & 0.01 & 0.03 & 0.064 \\
\addlinespace[2pt]
\midrule
\multicolumn{10}{@{}l}{\textit{ABC out-of-distribution set}} \\
Qwen3-VL-2B (fine-tuned) & 61.0
& \textbf{0.377} & \textbf{0.303} & \textbf{0.644} & \textbf{0.854}
& \textbf{0.10} & 0.032 & \textbf{0.11} & \textbf{0.29} \\
Qwen3-VL-2B (base) & 5.4
& 0.131 & 0.073 & 0.185 & 0.355
& 0.45 & 0.35 & 0.65 & 1.01 \\
GPT-5.2 High & \textbf{66.2}
& 0.344 & 0.289 & 0.509 & 0.730
& 0.22 & \textbf{0.03} & 0.12 & 0.33 \\
GPT-5.2 Medium & 62.6
& 0.346 & 0.285 & 0.557 & 0.747
& 0.22 & \textbf{0.03} & 0.13 & 0.34 \\
\bottomrule
\end{tabular}
\end{table}

\paragraph{In-Distribution Performance}
On Zero-to-CAD test data, the fine-tuned Qwen3-VL-2B-Instruct achieves an 82.1\% success rate with mean IoU of 0.747, substantially outperforming GPT-5.2 High (72.2\% success, 0.485 mean IoU). The base Qwen3-VL-2B-Instruct without fine-tuning achieves only a 6.6\% success rate, confirming that the task requires specialized training rather than relying on general vision-language capabilities. The median IoU of 0.847 and P90 of 0.999 indicate that successful reconstructions are typically geometrically accurate, with the top decile achieving near-perfect overlap.

\begin{figure}[h]
  \centering
  \includegraphics[width=0.6\linewidth]{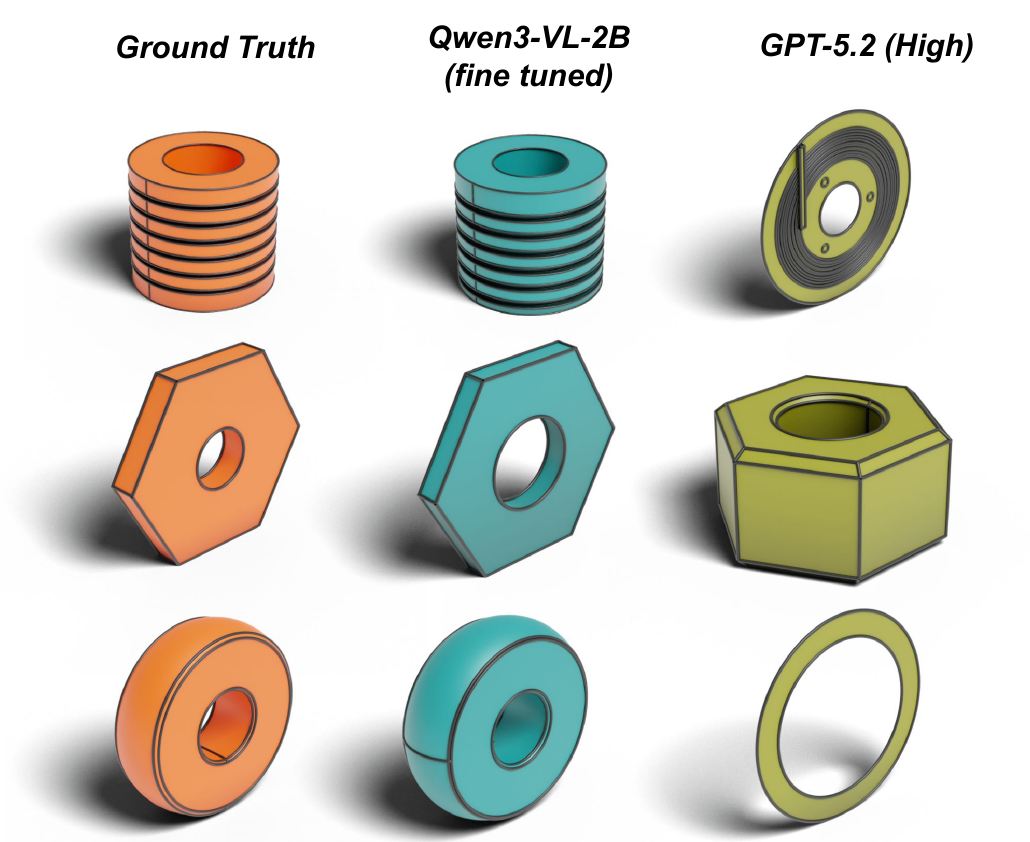}
  \caption{Qualitative comparison of Image-to-Sequence reconstruction on selected ABC samples, comparing ground truth, the fine-tuned Qwen3-VL-2B model, and GPT-5.2 outputs.}
  \label{fig:ABC_comparison}
\end{figure}

\paragraph{Out-of-Distribution Generalization}
On the ABC dataset, which contains human-designed CAD models with different stylistic conventions, the fine-tuned model maintains a 61.0\% success rate with mean IoU of 0.377. We chose ABC as the OOD benchmark because it consists of real-world human-designed CAD, making it a challenging benchmark for synthetic-to-real transfer; a performance drop relative to in-distribution evaluation is expected for all models. Nevertheless, the model generalizes meaningfully to real-world CAD data despite training exclusively on synthetic sequences. GPT-5.2 degrades less from Zero-to-CAD to ABC than the fine-tuned Qwen model, suggesting that synthetic-to-real transfer remains challenging despite strong in-distribution reconstruction. The fine-tuned model achieves higher IoU metrics (mean, median, P75, P90) than GPT-5.2 variants, though GPT-5.2 High achieves a slightly higher success rate (66.2\%) on this out-of-distribution test. This higher success rate comes at lower geometric fidelity: the fine-tuned 2B model outperforms GPT-5.2 on all IoU statistics and most CD statistics, while GPT-5.2 has a slightly better median CD. Figure~\ref{fig:ABC_comparison} illustrates representative cases where the fine-tuned model captures the essential geometry of ABC parts more faithfully than GPT-5.2.

The key takeaway is that bootstrapping is possible: a 2B model trained exclusively on synthetic Zero-to-CAD data achieves meaningful reconstruction of human-designed B-Rep geometries, directly supporting the research question of whether LLM priors about mechanical parts can be converted into useful construction-history supervision without any real sequence data.

\section{Conclusion}

We presented Zero-to-CAD, an agentic pipeline that synthesizes executable CAD construction sequences without relying on real-world design histories. By combining LLM generation with execution feedback, documentation lookup, and multi-stage validation, the pipeline produces geometrically valid, human-readable programs with named parameters and broad operation coverage, including Booleans, fillets, chamfers, shells, lofts, sweeps, and patterns. The resulting release contains 999,633 executable sequences, a curated subset of 100,000 with precomputed embeddings, the fine-tuned 2B vision-language model, system prompts, and inference code. Our Image-to-Sequence experiments show that models trained on this synthetic supervision can bootstrap editable CAD reconstruction, reaching 82.1\% success in-distribution and generalizing to human-designed ABC parts. These results suggest that CAD sequence modeling can progress through scalable synthesis, while leaving open broader questions around synthetic data provenance and attribution.

\bibliographystyle{tmlr}
\bibliography{main}

\clearpage
\appendix
\section*{Appendix}

\section{Dataset Samples}
\begin{figure*}[h]
\centering
\includegraphics[width=0.64\linewidth]{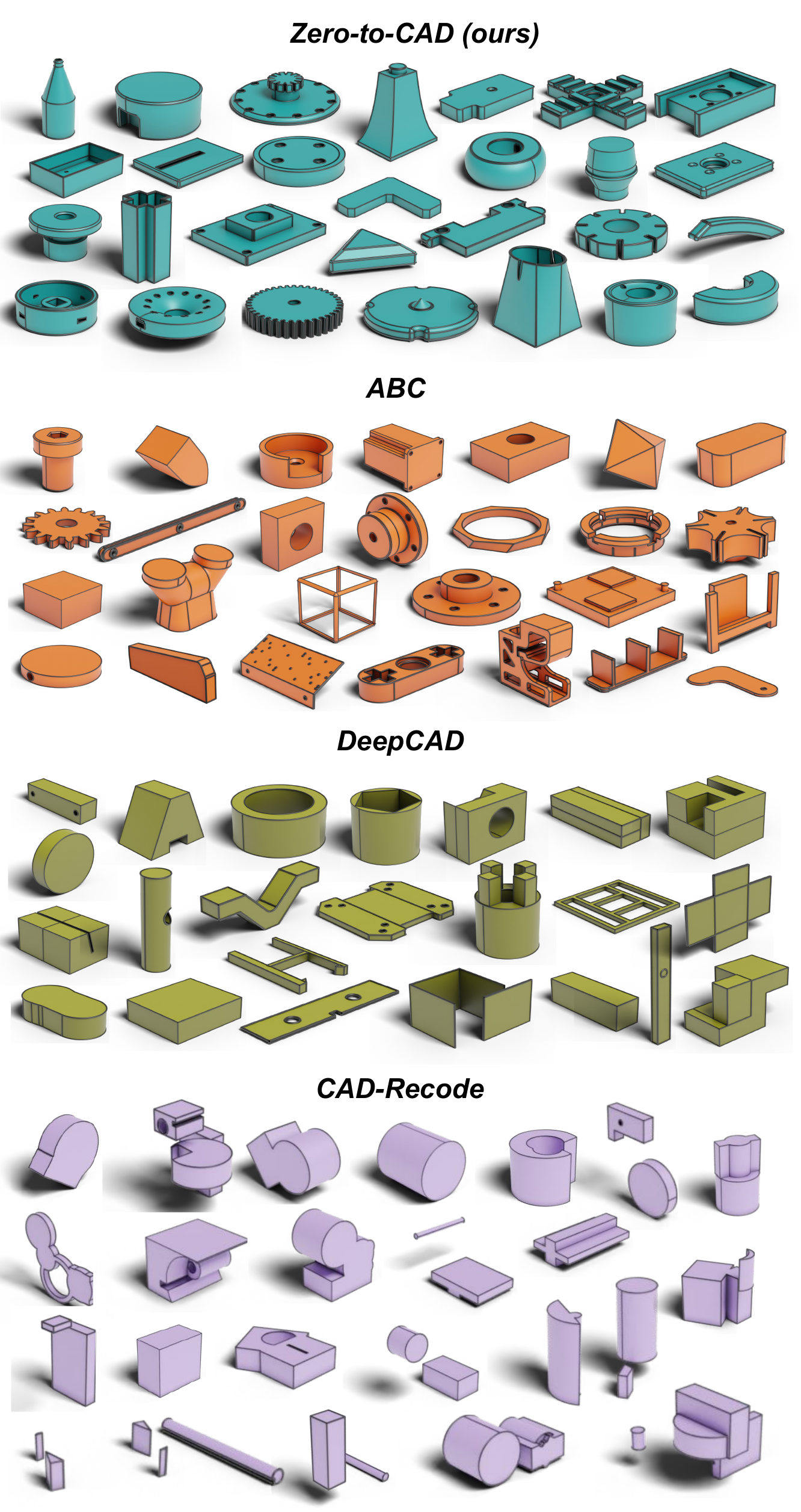}
\caption{Visual comparison of dataset samples from Zero-to-CAD, ABC, DeepCAD, and CAD-Recode.}
\label{fig:dataset_comparison}
\end{figure*}

\clearpage
\section{Generation Statistics Distributions}
\label{sec:generation_distributions}
\begin{figure*}[h]
\centering
\begin{subfigure}[b]{0.48\textwidth}
    \centering
    \includegraphics[width=\linewidth]{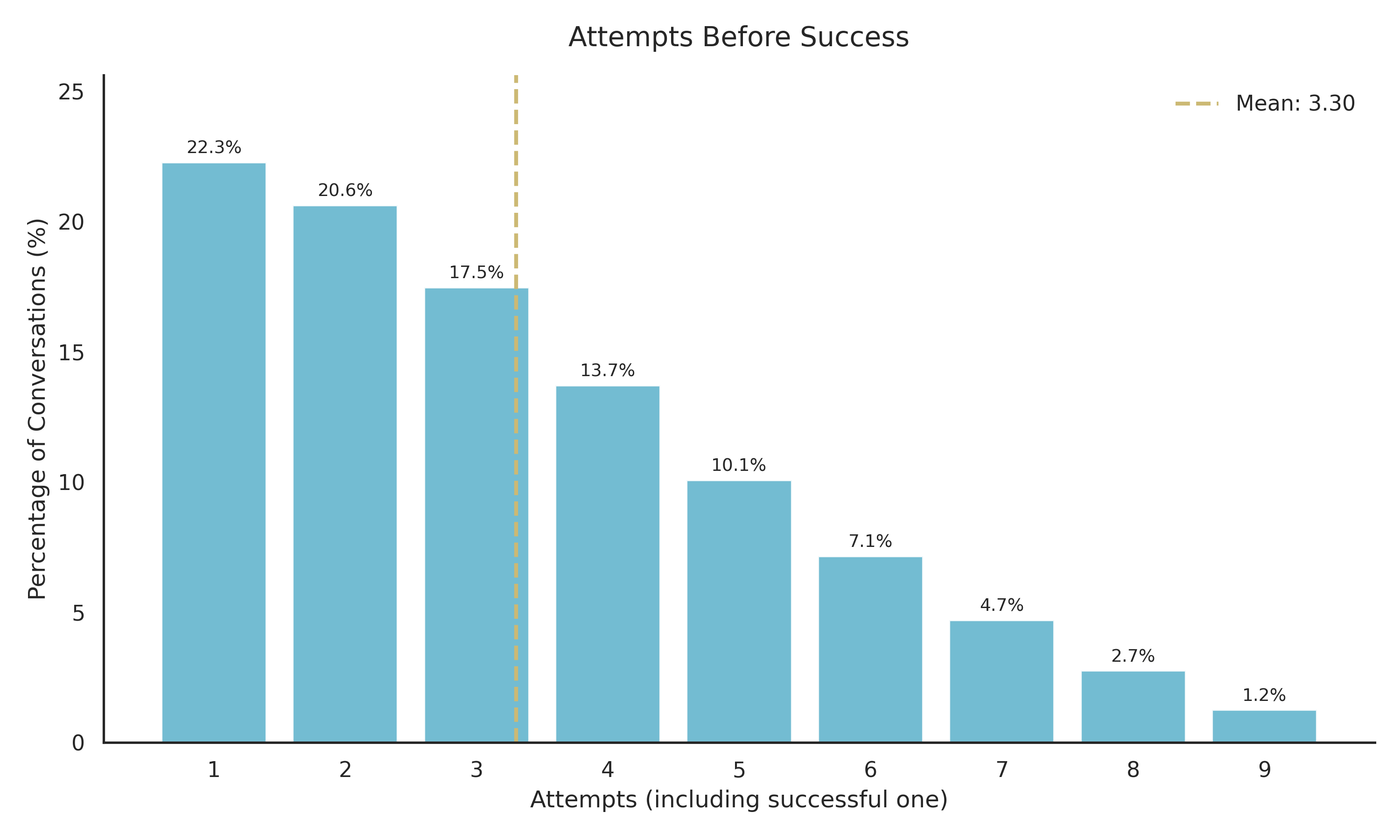}
    \caption{Validation attempts}
    \label{fig:attempts_dist}
\end{subfigure}
\hfill
\begin{subfigure}[b]{0.48\textwidth}
    \centering
    \includegraphics[width=\linewidth]{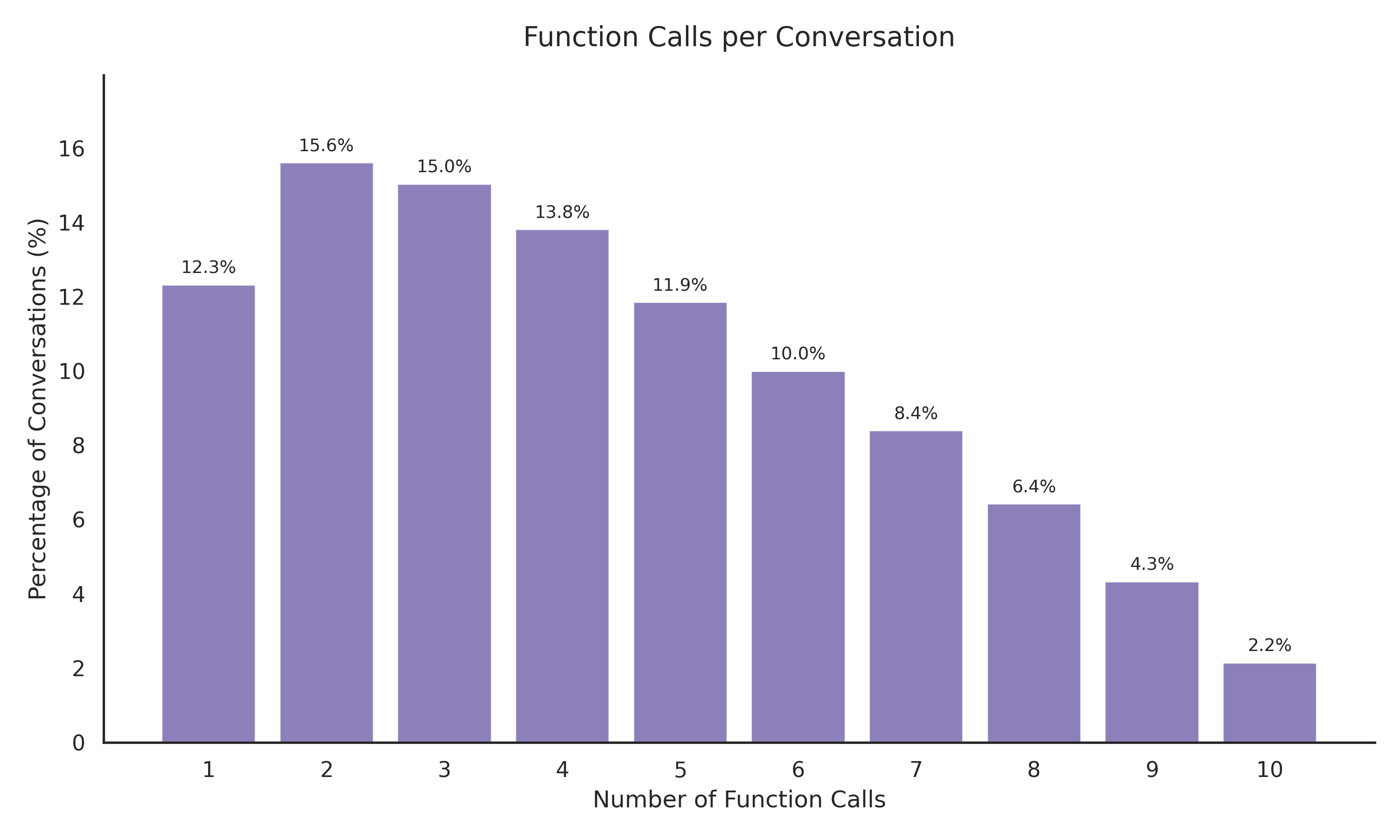}
    \caption{Function calls}
    \label{fig:function_calls_dist}
\end{subfigure}

\vspace{0.5em}

\begin{subfigure}[b]{0.48\textwidth}
    \centering
    \includegraphics[width=\linewidth]{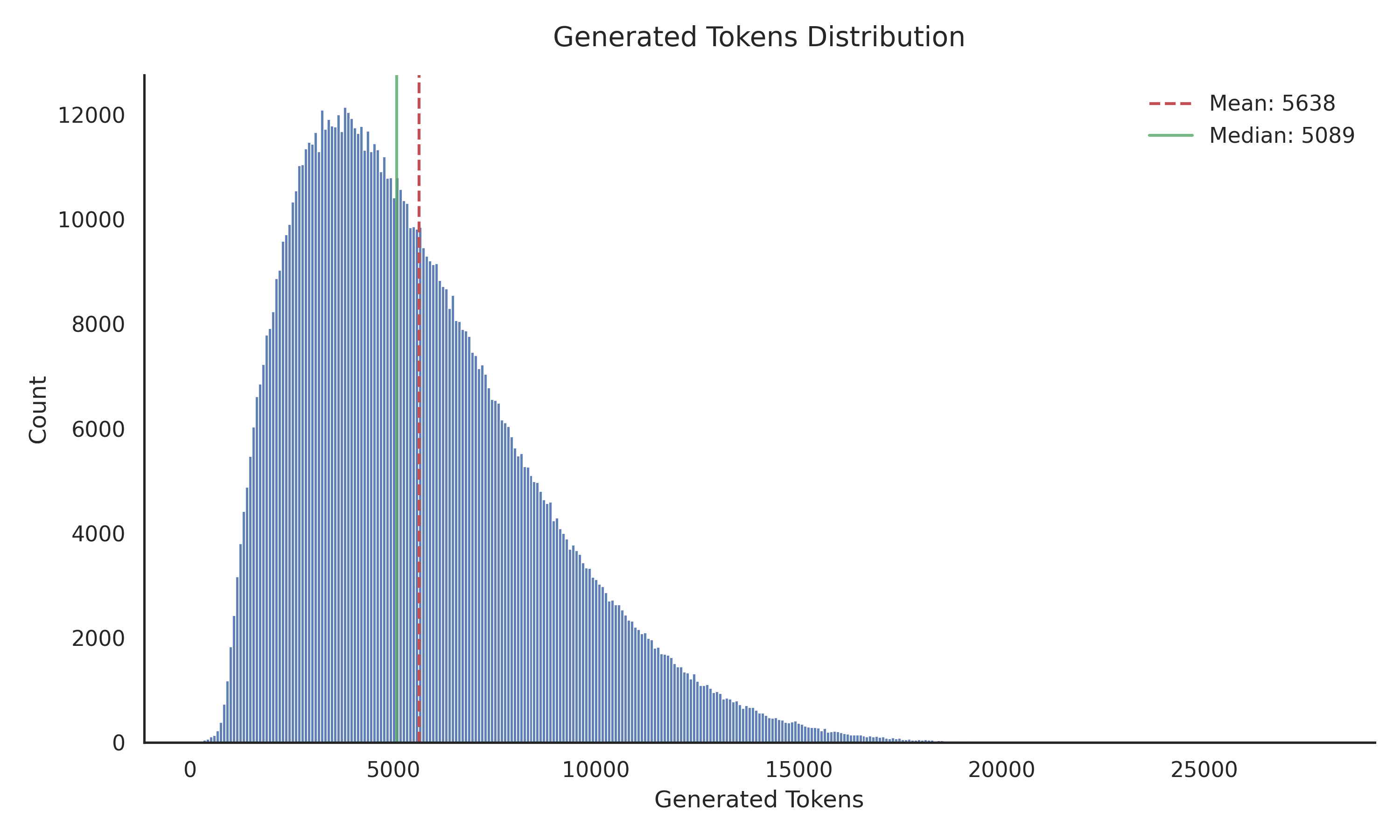}
    \caption{Generated tokens}
    \label{fig:tokens_dist}
\end{subfigure}
\hfill
\begin{subfigure}[b]{0.48\textwidth}
    \centering
    \includegraphics[width=\linewidth]{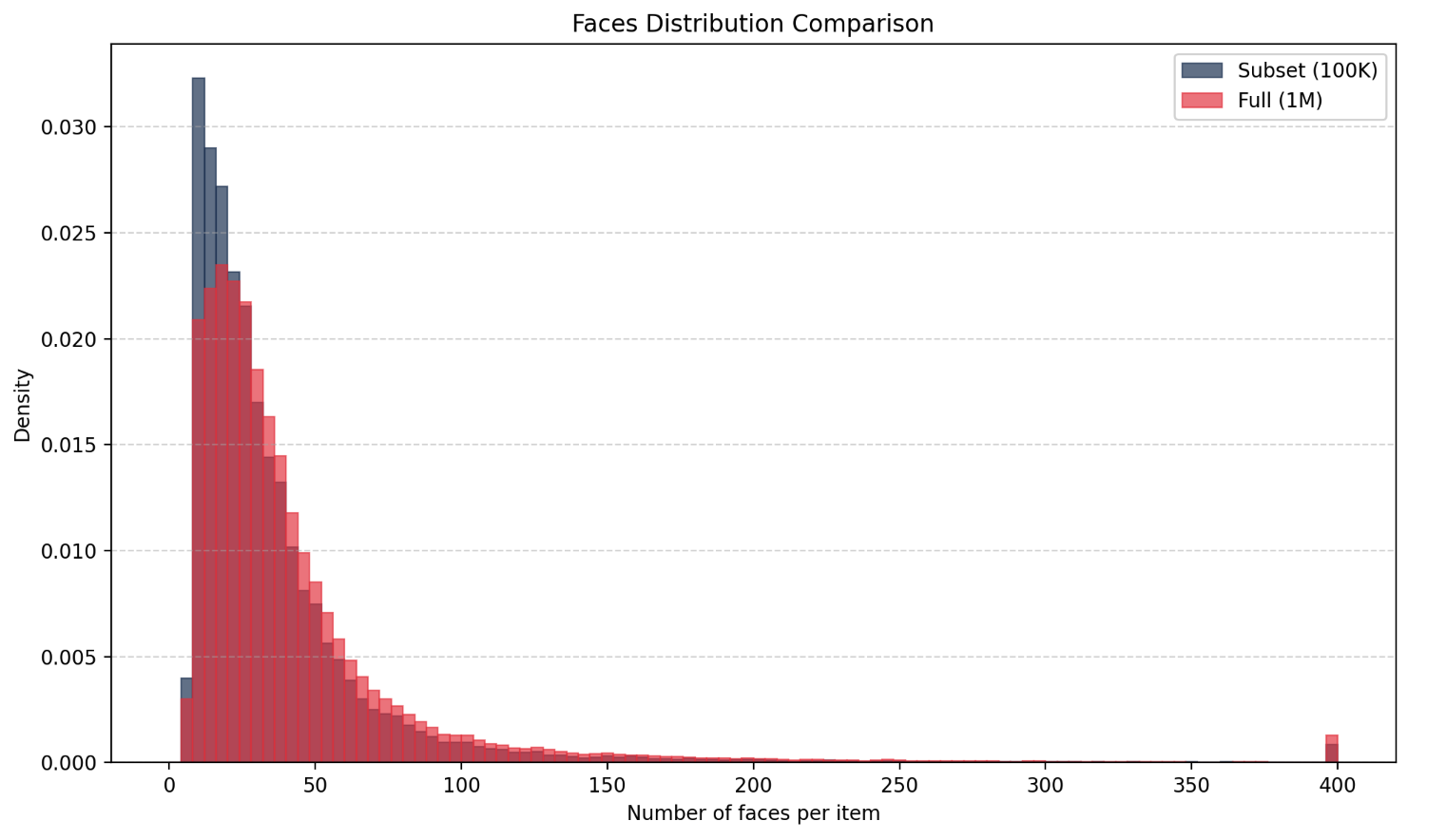}
    \caption{Face counts}
    \label{fig:face_count_dist}
\end{subfigure}

\vspace{0.5em}

\begin{subfigure}[b]{0.7\textwidth}
    \centering
    \includegraphics[width=\linewidth]{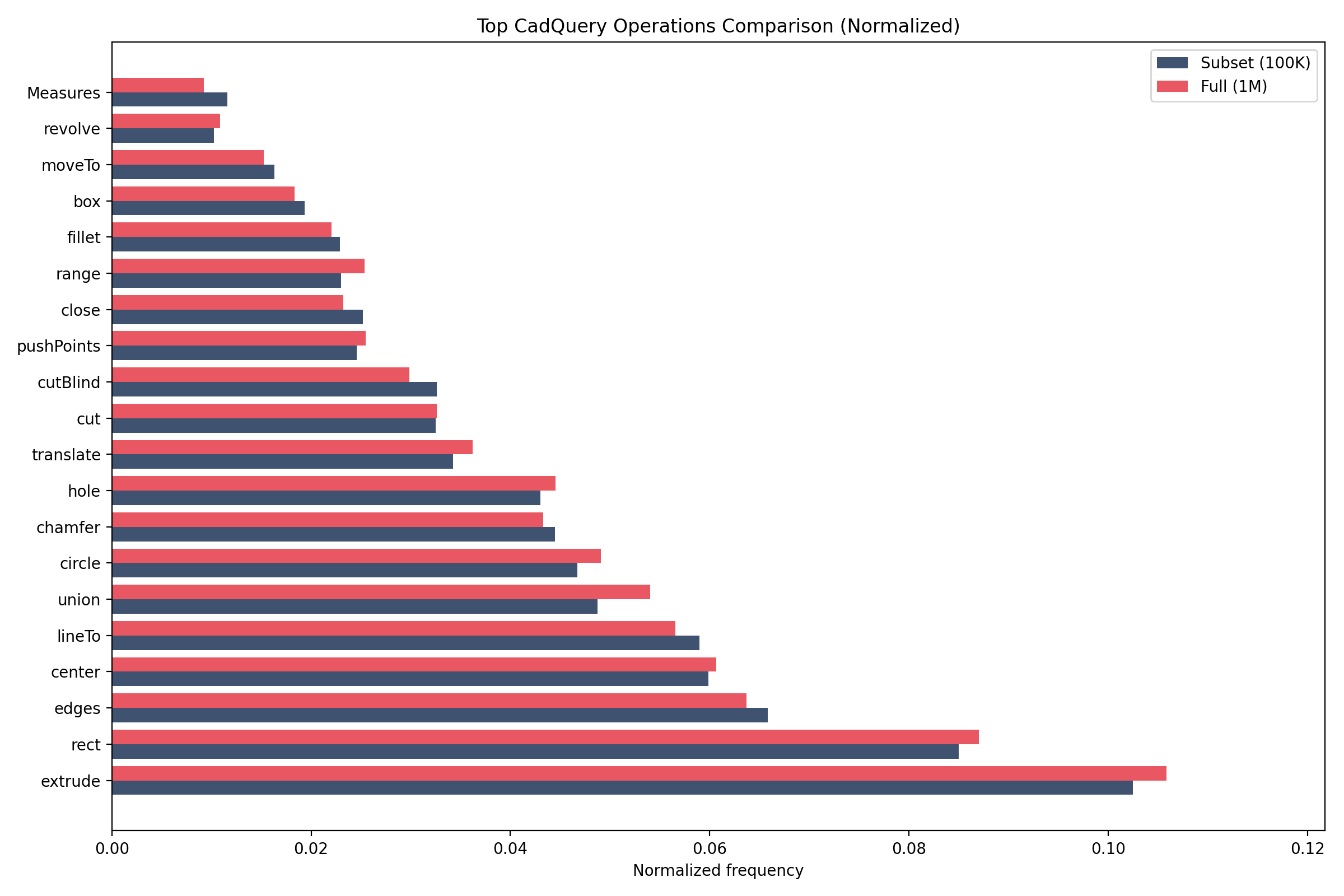}
    \caption{Operation coverage}
    \label{fig:ops_dist}
\end{subfigure}

\caption{Generation statistics: validation attempts before success, function calls per conversation, generated tokens per design, face counts, and CAD operation coverage.}
\label{fig:generation_distributions}
\end{figure*}

Figure~\ref{fig:generation_distributions} shows detailed distributions of the dataset generation process, including validation attempts, function call frequencies, token counts, geometric complexity (face counts), and operation coverage.

\clearpage

\section{Training Details}
\label{sec:training_details}

Table~\ref{tab:training_hyperparams} summarizes the hyperparameters used for fine-tuning the vision-language model on Zero-to-CAD data.

\begin{table}[h]
\caption{Fine-tuning configuration for Qwen3-VL-2B-Instruct on Zero-to-CAD.}
\label{tab:training_hyperparams}
\centering
\footnotesize
\setlength{\tabcolsep}{6pt}
\renewcommand{\arraystretch}{1.08}
\begin{tabular}{@{}l l@{}}
\toprule
\textbf{Hyperparameter} & \textbf{Value} \\
\midrule
\multicolumn{2}{@{}l}{\textit{Model setup}} \\
Base model & Qwen3-VL-2B-Instruct \\
Training mode & Full fine-tuning \\
Max sequence length & 4,096 tokens \\
Attention dropout & 0.1 \\
\addlinespace[2pt]
\multicolumn{2}{@{}l}{\textit{Optimization}} \\
Optimizer & AdamW \\
Base learning rate & $1 \times 10^{-4}$ \\
Vision tower learning rate & $1 \times 10^{-4}$ \\
MLP projector learning rate & $1 \times 10^{-4}$ \\
Weight decay & 0.0 \\
LR scheduler & Cosine \\
Warmup ratio & 0.03 \\
\addlinespace[2pt]
\multicolumn{2}{@{}l}{\textit{Distributed training}} \\
Number of GPUs & 16 H100 80GB GPUs \\
Per-GPU batch size & 1 \\
Gradient accumulation steps & 1 \\
Effective batch size & 16 \\
Number of epochs & 3 \\
Distributed strategy & DDP \\
Precision & bfloat16 \\
\bottomrule
\end{tabular}
\end{table}

\section{System Prompts}
\label{sec:system_prompts}

We provide the system prompts used in both the dataset generation pipeline and the downstream fine-tuning experiments.

\subsection{Catalog Generation Prompt}

The catalog generation stage uses a prompt (Figure~\ref{fig:catalog_prompt}) that instructs the LLM to act as an expert mechanical parts librarian. The prompt emphasizes producing concise, plausible descriptions without dimensions, ensuring uniqueness within each batch, and outputting results as a JSON array for programmatic processing.

\begin{figure*}[h]
\small
\centering
\fbox{\parbox{0.97\textwidth}{
\textbf{Catalog Generation Prompt}\\[0.5em]
You are an expert mechanical parts librarian.\\[0.3em]
Produce concise, one-sentence engineering part descriptions commonly seen in datasets like ABC. Requirements:\\
1. Each item is a single, self-contained part (not an assembly)\\
2. Each item is 1-3 sentences only, plain text (no numbering)\\
3. Be specific and plausible (e.g., ``flat plate bracket with 4 holes'')\\
4. Avoid speculative language or marketing terms\\
5. Ensure uniqueness within the batch (no duplicates or near-duplicates)\\
6. No need to specify the material of the part\\[0.3em]
DO NOT INCLUDE ANY DIMENSIONS IN THE DESCRIPTIONS, just the type and key features of the part.\\[0.3em]
Do not call any tools. Do not include explanations or code fences. Output only a JSON array of strings.
}}
\caption{System prompt for catalog description generation. The LLM generates batches of part descriptions that specify types and features without dimensions, enabling diverse yet semantically meaningful specifications.}
\label{fig:catalog_prompt}
\end{figure*}

\subsection{Code Generation Prompt}

The CAD code generation stage uses an extensive system prompt (Figure~\ref{fig:gen_prompt}) that encodes 19 design principles covering parametric design, CadQuery best practices, scale conventions, manufacturability constraints, and error-handling protocols. The prompt instructs the model to maintain geometric sophistication when debugging---looking up correct syntax rather than simplifying code---and provides guidance on tool usage for validation and documentation lookup.

\begin{figure*}[h]
\small
\centering
\fbox{\parbox{0.97\textwidth}{
\textbf{Code Generation Prompt (excerpt)}\\[0.5em]
You are an expert CAD engineer specialized in CadQuery, a Python-based parametric CAD library.\\[0.3em]
Your task is to generate clean, well-structured CadQuery code following these principles:\\
1. Always separate numerical variable definitions from operations\\
2. Use descriptive variable names\\
3. Do NOT add comments to the code\\
4. The final result must be stored in a variable called \texttt{result}\\
5. Never include export statements - exports are handled separately\\
6. Ensure all geometry is valid and manufacturable\\
7. Follow CadQuery best practices and syntax\\
8. SCALE CONVENTION: Use a maximum dimension of 100 units (treat 100 units as 10 cm in real-world scale)\\
9. SELF-CONTAINED CODE: Each code output must be completely self-contained and executable\\
10. For starting shapes, prefer constructing a plausible 2D sketch and then using extrude or revolve\\
11. For sketches, avoid trivial single-primitive profiles; build composite, non-trivial closed profiles\\
12. Make designs resemble plausible real-world components with clear intent (bracket, clamp, flange, etc.)\\
13. Anticipate future complexity: expose accessible faces for later sketches, maintain symmetry planes\\
14. Keep key dimensions as named variables to support later variation\\
15. Keep the part near the global origin with stable orientation\\
16. CRITICAL: Generate DETAILED, SOPHISTICATED code with rich geometric complexity\\
17. EDGE BREAKS: When appropriate, add small chamfers or fillets to break sharp edges\\
18. HOLE PLACEMENT: Choose mechanically sensible faces and locations aligned to datums\\
19. SYMMETRY: Prefer symmetric layouts; break symmetry only with clear functional justification\\[0.3em]
IMPORTANT: You have access to tools: execute\_and\_validate, lookup\_documentation, grep\_documentation\\[0.3em]
WHEN YOU ENCOUNTER AN ERROR: DO NOT simplify the code. Use documentation tools to find correct syntax. Fix the SPECIFIC error while maintaining all complexity. FORBIDDEN: Do not remove features or complexity to make errors go away.
}}
\caption{Excerpt of the generation system prompt used for CAD sequence synthesis. The full prompt includes detailed CadQuery API signatures and additional guidance on sketch construction, revolve operations, and error recovery protocols.}
\label{fig:gen_prompt}
\end{figure*}

\subsection{Inference System Prompts}

For the downstream Image-to-Sequence task, we use different system prompts depending on whether the model has been fine-tuned (Figure~\ref{fig:inference_prompts}). The fine-tuned Qwen model uses a minimal prompt, as it has internalized the task requirements during training. Zero-shot models (base Qwen and GPT-5.2) use a longer prompt with explicit instructions to store results in a specific variable and avoid export commands, since these models require guidance on output format. This difference reflects the distinction between a model trained on the task versus one prompted at inference time.

\begin{figure*}[h]
\small
\centering
\fbox{\parbox{0.47\textwidth}{
\textbf{Fine-tuned Model Prompt}\\[0.5em]
You are a CAD code assistant. Given multiple rendered views of a 3D shape, generate clean, well-structured CadQuery Python code that accurately reproduces the geometry.
}}
\hfill
\fbox{\parbox{0.47\textwidth}{
\textbf{Zero-shot Model Prompt}\\[0.5em]
You are a CAD code assistant. Given multiple rendered views of a 3D shape, generate clean, well-structured CadQuery Python code that accurately reproduces the geometry. Store the final shape representation in the \texttt{result = ...} variable. Do not add any export commands. Keep the shape in the \texttt{result} variable only, as the export code will be appended later.
}}
\caption{System prompts for Image-to-Sequence inference. The fine-tuned model prompt (left) is minimal since training has internalized task conventions. The zero-shot prompt (right), used for both base Qwen and GPT-5.2, includes explicit output format instructions.}
\label{fig:inference_prompts}
\end{figure*}

\section{Example Generated Code}
\label{sec:example_code}

Figure~\ref{fig:example_code} shows the complete CadQuery code for the mounting plate depicted in Figure~\ref{fig:zero_to_cad_architecture}. The code demonstrates several characteristics of Zero-to-CAD outputs: descriptive variable names (e.g., \texttt{plate\_thickness}, \texttt{fillet\_radius}), logical construction order (base plate, then rib, then subtractive features), and a mix of operations (extrusion, union, cut, fillet, chamfer). This interpretable structure allows engineers to modify dimensions or adapt the design for new requirements.

\begin{figure*}[h]
\centering
\begin{minipage}{0.98\textwidth}
\begin{lstlisting}[style=paperstyle, language=Python]
import cadquery as cq

plate_length = 60.0
plate_width = 40.0
plate_thickness = 8.0
rib_length = 20.0
rib_width = 6.0
hole_diameter = 8.0
mount_hole_dia = 3.0
mount_hole_offset = 5.0
fillet_radius = 2.0
chamfer_distance = 1.0
slot_width = 4.0
slot_length = 30.0

base = (
    cq.Workplane('XY')
    .rect(plate_length, plate_width, centered=True)
    .extrude(plate_thickness)
)
base = base.edges("|Z").fillet(fillet_radius)

rib = (
    cq.Workplane('XY')
    .center(-plate_length/2 + rib_length/2, 0)
    .rect(rib_length, rib_width, centered=True)
    .extrude(plate_thickness)
)
bracket = base.union(rib)

bracket = bracket.cut(
    cq.Workplane('XY').center(0, 0).circle(hole_diameter/2).extrude(plate_thickness + 2)
)

slot = (
    cq.Workplane('XY')
    .center(plate_length/2 - slot_length/2, 0)
    .rect(slot_length, slot_width, centered=True)
    .extrude(plate_thickness)
)
bracket = bracket.cut(slot)

mount_positions = [
    (-plate_length/2 + mount_hole_offset, -plate_width/2 + mount_hole_offset),
    (-plate_length/2 + mount_hole_offset, plate_width/2 - mount_hole_offset),
    (plate_length/2 - mount_hole_offset, -plate_width/2 + mount_hole_offset),
    (plate_length/2 - mount_hole_offset, plate_width/2 - mount_hole_offset),
]
for x, y in mount_positions:
    bracket = bracket.cut(
        cq.Workplane('XY').center(x, y).circle(mount_hole_dia/2).extrude(plate_thickness + 2)
    )

bracket = bracket.edges("|Z").chamfer(chamfer_distance)
result = bracket
\end{lstlisting}
\end{minipage}

\caption{Complete CadQuery code for the mounting plate shown in Figure~\ref{fig:zero_to_cad_architecture}. The code exhibits interpretable structure with named parameters, logical construction order, and diverse operations including extrusion, Boolean union/cut, fillet, and chamfer.}
\label{fig:example_code}
\end{figure*}

\end{document}